# The Last State of Artificial Intelligence in Project Management

Mohammad Reza Davahli[1*],

[1]Department of Industrial Engineering and Management Systems, University of Central Florida, Orlando, FL 32816, USA

*Correspondence: Mohammad Reza Davahli

E-mails: mohammadreza.davahli@ucf.edu (M.R. Davahli),

**Abstract:**

Artificial intelligence (AI) has been used to advance different fields, such as education, healthcare, and finance. However, the application of AI in the field of project management (PM) has not progressed equally. This paper reports on a systematic review of the published studies used to investigate the application of AI in PM. This systematic review identified relevant papers using Web of Science, Science Direct, and Google Scholar databases. Of the 652 articles found, 58 met the predefined criteria and were included in the review. Included papers were classified per the following dimensions: PM knowledge areas, PM processes, and AI techniques. The results indicated that the application of AI in PM was in its early stages and AI models have not applied for multiple PM processes especially in processes groups of project stakeholder management, project procurements management, and project communication management. However, the most popular PM processes among included papers were project effort prediction and cost estimation, and the most popular AI techniques were support vector machine, neural networks, and genetic algorithms.

# 1. Introduction

The rapid development of global economies has a significant impact on the number and complexity of projects [1]. Accordingly, increasing the complexity of projects makes project management a challenging and complicated task [1,2]. In recent years, there have been several attempts to ease the difficulty of project management by using advanced analytical techniques [1,3–7]. One of these techniques is artificial intelligence (AI) which has significant potential for improving the quality of project management (PM) [1,3–7].

From the beginning until the end of a project, a considerable amount of data is gathered and analyzed [8]. These data are collected from different processes of PM and are shared between project teams [8]. AI models can be trained by these data to assist project managers [9]. For example, it is reported that AI can decrease uncertainties in PM by using logical reasoning and probability calculation [6,10].

In this study, all academic publications that investigated the application of AI in different processes of PM are selected, categorized, and reported. For this purpose, a systematic review is used to identify, summarize, and analyze the findings of all relevant individual studies that are addressing predefined research questions. This review is structured as follows: the methodology section discusses inclusion and exclusion criteria and the risk of bias; the results section provides outputs of the literature search; the discussion section describes the current level of AI application in PM.

# 2. Methodology

The Preferred Reporting Items for Systematic Reviews and Meta-Analyses (PRISMA) guidelines were followed for this systematic review [11] and two main features of research questions and search strategies were developed. The following research questions have guided this review:

RQ1. How can research on the application of AI in PM be classified?

RQ2. What AI techniques have been applied to PM processes?

To answer research questions, a search strategy was developed by (1) defining keywords and identifying all relevant records; (2) filtering the identified articles; and (3) addressing the risk of bias among records [12,13]. Two sets of keywords were defined and a combination of the first set and the second set was used to identify relevant papers.

- First set: Artificial intelligence, pattern recognition, machine learning, deep learning, convolutional neural networks, and computational intelligence.
- Second set: Project management, project integration management, project scope management, project schedule management, project cost management, project quality management, project resource management, project communications management, project risk management, project procurement management, and project stakeholder management.

Web of Science, Science Direct, and Google Scholar were used as a database to identify relevant materials. After identifying 652 articles with relevant content, a formal screening process was followed based on inclusion and exclusion criteria. The inclusion criteria were articles related to project management and artificial intelligence, articles written in English, articles published from 2000 to 2019, and articles related to research questions. The exclusion criteria were papers written in other languages, chapters of books, letters, newspaper articles, viewpoints, presentations, anecdotes, duplicated studies, short papers, and posters, and articles from secondary sources that were not free or open access.

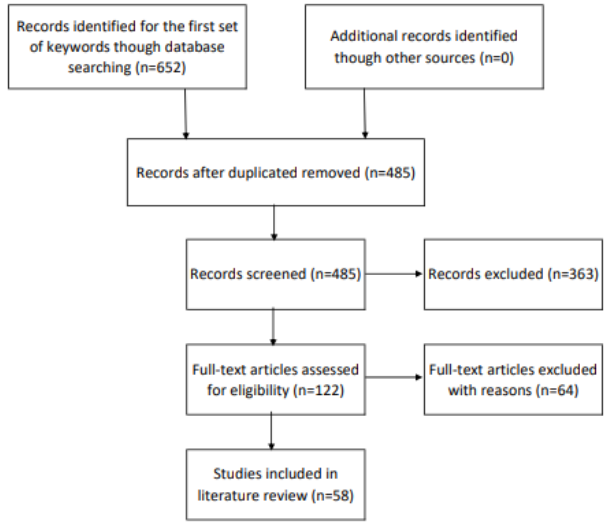

Figure 1. Chart of selection strategy following PRISMA guidelines.

Selection bias in a systematic review can occur by (1) applying inclusion/ exclusion criteria, and/or (2) categorizing included papers into relevant dimensions. To address the first type of bias, the author and his assistant independently reviewed the abstract and conclusion sections of records and selected articles for full-text review [14]. After comparing their selected articles, they independently read the full-text of selected articles and decided whether to include the article or not. Subsequently, the author and his assistant compared their lists to reach a unified list. To address the second type of bias, they separately specified relevant processes and areas of included papers. Subsequently, they compared the results and resolved disagreements. The chart of selection strategy based on the PRISMA guidelines is shown in Figure1. As a result, 58 papers were included in this literature review.

## 3. Results

List of included papers accompanied by PM process groups, PM Knowledge areas, and AI techniques are represented in Table 1.

Table 1. Included papers in the application of AI in PM.

| Reference | PM Process group | PM Knowledge area | PM process | AI technique |
|---|---|---|---|---|
| [15] | Executing | Communication | Documents classification | Support vector machine (SVM) |
| [16] | Planning | Cost | Cost estimation | Fuzzy decision trees |
| [17] | Planning | Cost | Cost estimation | Linear regression, support vector regression (SVR), and multilayer perceptron (MLP) |
| [18] | Planning | Cost | Cost estimation | Genetic algorithm |
| [4] | Monitoring | Cost | Cash flows | K-means clustering and evolutionary fuzzy neural inference model (EFNIM) |
| [19] | Monitoring | Cost | Estimate at Completion | Evolutionary support vector machine inference model (ESIM) |
| [5] | Monitoring | Cost | Estimation at completion | Global harmony search (GHS) and brute force (BF) integrated with extreme learning machine (ELM) |
| [20] | Monitoring | Cost | Cash flow control | Evolutionary fuzzy hybrid neural network (EFHNN) |
| [21] | Planning | Cost | Construction cost | A fusion of individual multilayer perceptron neural networks |
| [22] | Monitoring | Cost | Construction Cost Index (CCI) | Self-adaptive structural radial basis neural network intelligence machine (SSRIM) |
| [23] | Monitoring | Cost | Construction site overhead costs | A fusion of individual multilayer perceptron neural networks |
| [24] | Planning | Cost | Construction costs | A combination of genetic algorithm and artificial neural networks (GA-ANN) |
| [25] | Planning | Cost | Site overhead costs | Artificial neural networks |
| [26] | Planning | Cost | Cost of risk | Bayesian networks |
| [27] | Planning | Cost | Construction cost | Unsupervised deep Boltzmann machine (DBM) learning approach along with a softmax layer (DBM-softmax), and a three-layer backpropagation neural network (BPNN) or another regression model, and support vector machine (SVM) |
| [28] | Monitoring | Cost | Earned value and earned schedule | Support vector machines |

| Reference | PM Process group | PM Knowledge area | PM process | AI technique |
|---|---|---|---|---|
| [29] | Monitoring | Cost | Cash flow control | Evolutionary fuzzy hybrid neural network (EFHNN) |
| [30] | Executing | Resource | Project Manager competencies | ANN-based framework named as I-Competere |
| [31] | Monitoring | Integration | Decision making | Evolutionary support vector machine inference system |
| [32] | Monitoring | Integration | Decision making | Evolutionary fuzzy support vector machine inference model (EFSIM) |
| [33] | Monitoring | Integration | Project dispute problems | Fast messy genetic algorithm (fmGA) - based SVM (GASVM) |
| [34] | Monitoring | Integration | Project dispute problems | Multilayer perceptron (MLP) neural networks, decision trees (DTS), support vector machines, the naive Bayes classifier, and K-nearest neighbor |
| [35] | Monitoring | Integration | Decision making | Fuzzy clustering-based genetic algorithm (FCGA) approach |
| [36] | Initiating | Integration | Effort prediction | A Treeboost (Stochastic Gradient Boosting) model |
| [37] | Initiating | Integration | Effort prediction | Decision tree learning algorithm |
| [38] | Initiating | Integration | Software project effort prediction | A framework based on fuzzy logic and neural networks |
| [39] | Initiating | Integration | Project success | Artificial neural networks |
| [40] | Initiating | Integration | Effort estimation | Random forest |
| [41] | Initiating | Integration | Effort estimation | Bootstrap-based Neural Networks |
| [42] | Initiating | Integration | Effort estimation | Support vector machines, neural networks, and generalized linear models |
| [43] | Initiating | Integration | Project success | Evolutionary project success prediction model (EPSPM) |
| [44] | Initiating | Integration | Effort prediction | Artificial neural network, simple linear regression and multiple regressions |
| [45] | Initiating | Integration | Project success | Artificial neural networks ensemble and support vector machines classification models |
| [46] | Initiating | Integration | Project performance | Backpropagation deep belief network (BP-DBN) |
| [47] | Initiating | Integration | Project success | Artificial neural network (ANN) |

| Reference | PM Process group | PM Knowledge area | PM process | AI technique |
|---|---|---|---|---|
| [48] | Initiating | Integration | Project success | Evolutionary support vector machine inference model (ESIM) |
| [49] | Initiating | Integration | Effort prediction | Deep learning algorithms |
| [50] | Initiating | Integration | Effort prediction | Regression tree (M5P), multilayer perception (MLP) and support vector regression (SVR) |
| [51] | Monitoring | Integration | Project dispute resolution (PDR) | Fast messy genetic algorithm (fmGA) - based SVM (GASVM) |
| [52] | Monitoring | Quality | Quality of construction projects | Combination of Unascertained system and artificial neural network (ANN) |
| [53] | Monitoring | Schedule | Project scheduling | Metaheuristic tabu search algorithm |
| [54] | Planning | Risk | Risk management | Artificial neural networks (ANNS) |
| [55] | Planning | Risk | Risks assessment | Bayesian networks classifiers |
| [56] | Planning | Risk | Risk management and risk causality analysis | Bayesian networks (BNS) with causality constraints (BNCC) |
| [57] | Planning | Risk | Risk allocation decision-making process | Artificial neural network (ANN) |
| [58] | Planning | Scope | Requirements prioritization | Case-Based Ranking (CBRank) method |
| [59] | Planning | Scope | Stakeholder requirements | Naive Bayes and the M5P-Tree algorithms |
| [60] | Planning | Schedule | Resource constraint project scheduling problem (RCPSP) | The rout algorithm of reinforcement learning and support vector machine (SVM) |
| [61] | Planning | Schedule | Uncertainties at different levels of planning | Greedy algorithm and genetic algorithm |
| [62] | Planning | Schedule | Beta distributions for modeling activity duration | The artificial neural network-based approach |
| [63] | Planning | Schedule | Estimating the diaphragm wall duration | Firefly-tuned least squares support vector machine (FLSVM) |
| [64] | Monitoring | Schedule | Project delay | Decision tree and naïve Bayesian classification |
| [65] | Monitoring | Schedule | Schedule to completion (ESTC) | Neural network–long short-term memory (NN-LSTM) model |

| Reference | PM Process group | PM Knowledge area | PM process | AI technique |
|---|---|---|---|---|
| [66] | Planning | Schedule | Project planning and scheduling | Artificial neural networks and the fuzzy neural system |
| [67] | Planning | Schedule | The life cycle of new construction projects | Electromagnetism-like algorithm (EM) combined with K-nearest neighbor |
| [68] | Planning | Schedule | project planning recommendation system | Revised case-based reasoning (RCBR) algorithm |
| [7] | Monitoring | Schedule | Project duration and EVM | Decision tree, bagging, random forest, boosting, and support vector machine |
| [69] | Planning | Schedule | Real duration of a project | A nearest neighbor based extension model |

Project management can be categorized into five process groups, ten knowledge areas, and 49 processes as it is shown in Table 2 [8]. The PM process group is defined as *a logical grouping of PM processes to achieve objectives which are independent of project phases* [8]. The PM knowledge area is defined as *an identified area of PM defined by its knowledge requirements and described in terms of its component processes, practices, inputs, outputs, tools, and techniques* [8].

Table 2. Matrix of process groups and knowledge areas [8]

| knowledge Areas | Project Management Process Groups | | | | |
|---|---|---|---|---|---|
| | Initiating | Planning | Executing | Monitoring and Controlling | Closing |
| 4. Project Integration Management | 4.1. Develop Project Charter | 4.2. Develop Project Management Plan | 4.3. Direct and Manage Project Work<br>4.4 Manage Project Knowledge | 4.5. Monitor and Control Project Work<br><br>4.6 Perform Integrated Change Control | 4.7. Close Project or Phase |
| 5. Project Scope Management | | 5.1. Plan Scope Management<br>5.2. Collect Requirements<br>5.3. Define Scope<br>5.4. Create WBS | | 5.5. Validate Scope<br><br>5.6. Control Scope | |
| 6. Project Schedule Management | | 6.1. Plan Schedule Management<br>6.2. Define Activities<br>6.3. Sequence Activities<br>6.4. Estimate Activity Durations<br>6.5. Develop Schedule | | 6.6. Control Schedule | |
| | | 7.1. Plan Cost Management | | 7.4. Control Costs | |

| | | | | |
|---|---|---|---|---|
| 7. Project Cost Management | | 7.2. Estimate Costs 7.3. Determine Budget | | |
| 8. Project Quality Management | | 8.1. Plan Quality Management | 8.2. Manage Quality | 8.3. Control Quality |
| 9. Project Resource Management | | 9.1. Plan Resource Management 9.2. Estimate Activity Resources | 9.3. Acquired Resources 9.4. Develop Team 9.5. Manage Team | 9.6. Control Resources |
| 10. Project Communications Management | | 10.1. Plan Communications Management | 10.2. Manage Communications | 10.3. Monitor Communications |
| 11. Project Risk Management | | 11.1. Plan Risk Management 11.2. Identify Risks 11.3. Perform Qualitative Risk Analysis 11.4. Perform Quantitative Risk Analysis 11.5. Plan Risk Responses | 11.6. Implement Risk Responses | 11.7. Monitor Risks |
| 12. Project Procurement Management | | 12.1. Plan Procurement Management | 12.2. Conduct Procurements | 12.3. Control Procurements |
| 13. Project Stakeholder Management | 13.1. Identify Stakeholders | 13.2. Plan Stakeholder Engagement | 13.3. Manage Stakeholder Engagement | 13.4. Monitor Stakeholder Engagement |

Because many organizations use project-based structure, the effective project management has a significant impact on the growth of the economy [70,71]. Despite the importance of projects, PM methodologies have not adequately advanced, and these methodologies are surrounded by procedural and bureaucratic constraints [72]. As a response to the limitation of current PM methodologies, multiple scholars have attempted to use advanced analytical tools in project management [4,7]. Even though these attempts have not been practically successful, but recent AI advancements have created an opportunity to improve the quality of PM methodologies.

Artificial intelligence is described as *the ability of a machine to mimic intelligent human behavior, thus seeking to use human-inspired algorithms for approximating conventionally defiant problems* [73]. The main subfields of AI are pattern recognition (PR), machine learning (ML), and deep learning (DL) [10].

In pattern recognition, objects such as image, speech, and handwriting are classified into different categories and classes. PR does not focus on learning, and it mainly recognizes the patterns and classifies data [74,75]. In general, pattern recognition and machine learning are related topics, and they overlap in their areas [73,74]. However, ML is about training algorithms for learning, while PR focuses on classification methods [73,74].

Machine learning is defined as a *computational method using known properties of training data to make an accurate prediction or improve performance* [10]. ML can be categorized into supervised learning, reinforcement learning, semi-supervised learning, and unsupervised learning [73]. Supervised learning, in which a set of labeled data are used to make predictions and learn about unseen points, is associated with regression and classification [10,76]. Unsupervised learning predicts based on unlabeled data, and it is associated with dimensionality reduction and clustering [10,76]. Semi-supervised learning makes a prediction based on both unlabeled and labeled sampling data, and it is used in situations in which obtaining labeled data is expensive or unlabeled data are effortlessly accessible [10,76]. Reinforcement learning is about taking actions in the interacted environment to maximize rewards [10]. One of the main subfields of ML is deep learning, which is about learning and discovering complicated and unseen relationships among data [77].

## 4. Discussion

Caldas and associates (2002) is the only paper in the knowledge area of project communication management [15]. In this paper, the growing utilization of information technologies and the availability of electronic documents motivated the authors to study the automated document classification in construction projects [15]. Caldas and associates (2002) analyzed different text classification algorithms and they reported that the support vector machine algorithm had the best results [15]. The authors concluded that the main challenge in text classification of construction documents is having many classes and a small number of documents per class [15].

Out of 58, 16 papers focused on project cost management. Papatheocharous & Andreou, (2012) used fuzzy decision trees, as robust decision tree structures enhanced by fuzzy logic, to analyze project cost estimation. The study indicated that the proposed approach accurately estimated cost [16]. In this mater, Li and associates (2007) reported the effectiveness of the genetic algorithm for cost estimation [18]. In another study, Cheng and associates (2009) proposed an approach for valid predictions of project cash flows by using a K-means clustering and an evolutionary fuzzy neural inference model that employs a fuzzy logic, genetic algorithm, and neural network [4]. The K-means clustering is used for the categorization of similar construction projects and defining project properties, and the fuzzy logic and neural network are used for knowledge mapping and addressing uncertainties [4,29]. In one study, estimation of the project cash flow was investigated by developing an evolutionary fuzzy hybrid neural network that employs a hybrid neural network, fuzzy logic and genetic algorithm [20]. The result indicated the effective estimation of project cash flow in comparison with a neural network [20].

Two included papers focused on the estimate at completion (EAC). Cheng and associates (2010) investigated EAC, as an indicator of the final project cost, by using an evolutionary support vector machine inference model that combines a support vector machine and a fast messy genetic algorithm [19]. The authors concluded that estimate at completion using the evolutionary support vector machine inference model can accurately identify influential factors of the project cost [19]. Another study investigated generating a valid trend of EAC to help project managers to control the cost of projects [5]. In this study, two models of (1) global harmony search integrated with extreme learning machine, and (2) brute force integrated with extreme learning machine were developed [5]. The results indicated the effectiveness of both models in comparison with a benchmark model of artificial neural network [5]. The global harmony search optimization algorithm is an approach for optimizing a problem with discrete and continuous variables and the brute force input optimization method is an approach for solving problems and achieving solutions by checking the reliability of each option to satisfy the problems statement [5].

Three studies forecasted construction costs and overhead costs by developing a fusion of neural networks [21,23,25]. The results indicated the efficiency of developed models in comparison with single neural networks [21]. In another study, Cao and associates (2014) forecasted the variability of construction cost index (CCI) by developing a hybrid model of self-adaptive structural radial basis neural network intelligence machine (SSRIM) [22]. SSRIM employs multivariate adaptive regression splines (MARS), radial basis function neural network (RBFNN), and artificial bee colony (ABC) algorithm [22]. MARS is used to distinguish the importance and value of each input parameter in the forecasting model, and a combination of RBFNN and ABC is used to find optimal solutions [22].

Chou and associates (2015) forecasted construction costs by using different AI methods [24]. The results indicated that a combination of genetic algorithm and artificial neural networks (GA-ANN) had the most efficient performance [24]. In another study, Rafiei & Adeli (2018) investigated construction cost estimation by developing a novel machine learning as a fusion of unsupervised deep Boltzmann machine learning combined with softmax layer (DBM-SoftMax), backpropagation neural network (BPNN), and support vector machine [27]. In this study, DBM-SoftMax was used to extract features, and BPNN and support vector machine were used to create a supervised regression network [27]. The authors demonstrated that the proposed model performed more efficiently than the BPNN-only and support vector machine -only models [27].

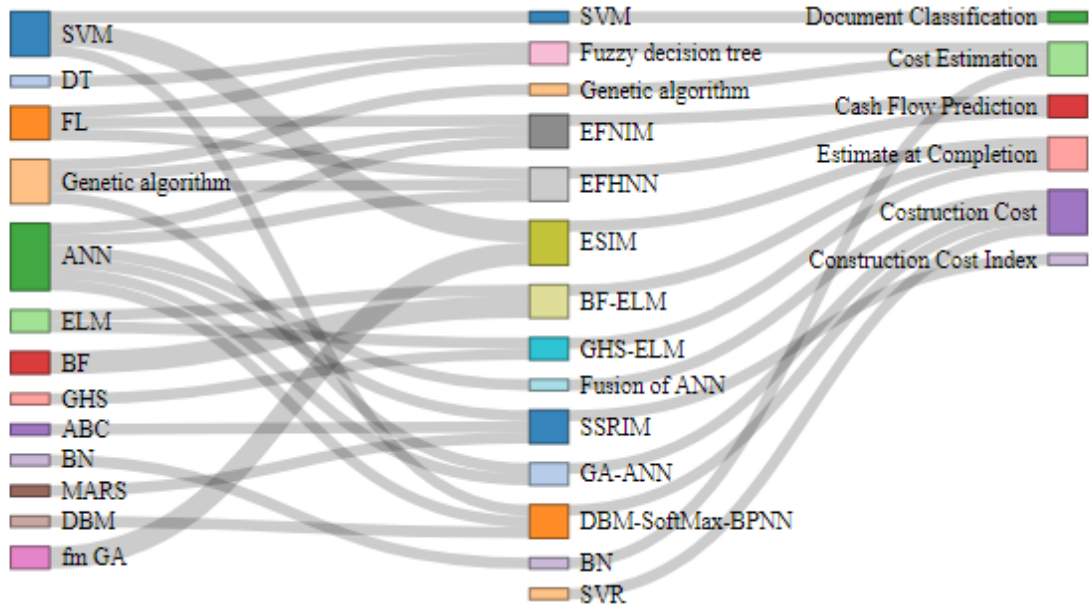

Figure. 2. Sankey diagram among single algorithms (left part), final AI models (middle section) and targeted process of project cost management (right part)

Khodakarami & Abdi (2014) investigated uncertainty of project cost by proposing a framework of Bayesian networks [26]. Bayesian networks were used to establish conditional dependency and causal relationships among different cost items [26]. The authors reported that it is also feasible to use the proposed model for scheduling the uncertainty of projects [26]. In another study, Wauters & Vanhoucke (2014) used a support vector regression model for the project time and cost forecasting analysis and compared the model with the most efficient earned schedule and earned value methods [28]. The authors reported that the proposed model performed better than the other methods [28]. The Sankey diagram of the application of different AI techniques in project cost management is represented in Figure 2.

Out of 58, 21 papers focused on project integration management. Wazirali and associates (2014) developed an evolutionary support vector machine inference system (ESIM) as a fusion of a support vector machine and a fast messy genetic algorithm to facilitate decision making by forecasting construction uncertainties [31]. In another study, Mungle and associates (2013) developed a fuzzy clustering-based genetic algorithm (FCGA) approach to facilitate decision making by addressing time–cost–quality trade-off problems [35]. The results indicated that the proposed model outperformed other multi-objective optimization methods [35]. In one study, Cheng & Roy (2010) developed an evolutionary fuzzy support vector machine inference model (EFSIM) as a fusion of a support vector machine, a fuzzy logic, and a fast messy genetic algorithm to facilitate decision making by retaining and utilizing experimental knowledge [32]. In this hybrid AI model, the fuzzy logic was deployed to handle approximate and vagueness reasoning, the fast messy genetic algorithm was used as an optimization tool, and the support vector machine was used to address fuzzy input-output mapping [32]. Chou and associates (2013, 2014) developed a fast messy genetic algorithm -based support vector machine (GASVM) as an integration of a support vector machine with a fast messy genetic algorithm to facilitate decision making by forecasting of the dispute propensity in projects [33,51]. In the proposed hybrid model, the fast messy genetic algorithm optimized the support vector machine parameters, and the support vector machine was used for learning and curve fitting [33].

One study focused on predicting project disputes, and it used hybrid and single classification techniques to develop prediction models [34]. In this study, the single classification methods are decision trees (DTs), multilayer perceptron (MLP) neural networks, a naïve Bayes classifier, a support vector machine, and a k-nearest neighbor [34]. The hybrid classification models are including (1) combination of clustering and classification techniques, and (2) combination of multiple classification techniques (DT+DT and MLP+MLP models) [34]. The results indicated that the hybrid models performed better than the single models, and also the multiple classification techniques outperformed the combination of clustering and classification techniques [34].

Nassif and associates (2012) explored the effort prediction in a software project by using a Treeboost (Stochastic Gradient Boosting) model [36]. The results indicated that the proposed model performed better than a multiple linear regression model [36]. In another study, Twala & Cartwright (2010) investigated the effort prediction in software projects by developing ensemble classifiers based on decision tree learning algorithms [37]. Another study addressed the effort prediction in software development projects by presenting SEffEst, which is a framework based on neural networks and a fuzzy logic to improve the accuracy of the effort estimation [38]. Mustapha & Abdelwahed (2019) investigated the effort estimation by using a random forest (RF) and comparing it with a classical regression tree [40]. The result indicated that the random forest outperformed the classical regression tree [40]. In one study, Braga and associates (2007) used a bagging technique to reduce the variance of prediction in three machine learning methods of a regression tree (M5P), a multilayer perception, and a support vector regression to estimate the project effort [50]. Tamura and associates (2018) studied the effort estimation in an open source software (OSS) project by developing a deep learning algorithm [49]. One study used artificial neural networks, a simple linear regression, and multiple regressions to estimate the effort of software projects [44]. The authors indicated that artificial neural networks outperformed other models in terms of accuracy of the effort estimation [44]

Mossalam & Arafa (2018) studied the project selection by analyzing project success factors and using a neural networks model [39]. Ko & Cheng (2007) developed an evolutionary project success prediction model as a fusion of fuzzy logic, genetic algorithms, and neural networks to distinguish and select project success factors [43]. In the proposed model, genetic algorithms were deployed for optimization, neural networks for finding the patterns between the inputs and the outputs and fuzzy logic for the approximate reasoning [43]. One study investigated project success factors by developing an evolutionary support vector machine inference model as a fusion of a support vector machine and a fast messy genetic algorithm [48]. Costantino and associates (2015) used neural networks for selecting project critical success factors to help project managers for evaluating projects in the selection phase [47]. Wang and associates (2012) used different methods for predicting project success factors in terms of the construction cost and schedule [45]. The authors indicated that the support vector machine outperformed other models in terms of predicting final project outcomes [45]. Sankey diagram of the application of different AI techniques in the project integration management is represented in Figure 3.

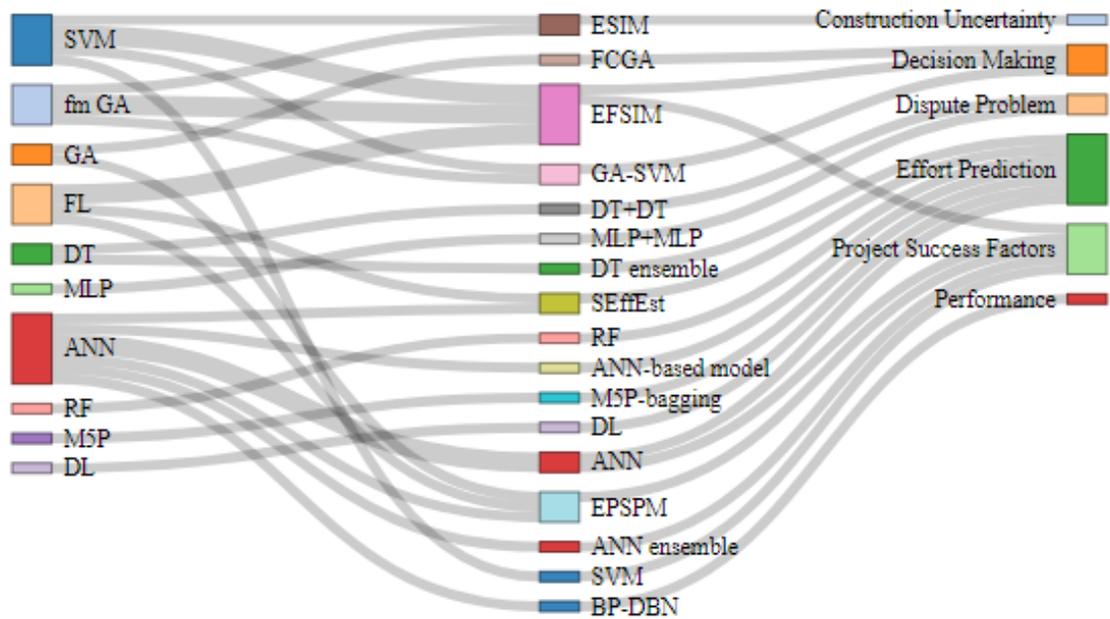

Figure 3. Sankey diagram among single algorithms (left part), final model (middle part) and process of project integration management (right part)

In the project resource management, one study used an artificial neural network-based framework, namely, I-Competere, in the human resource management to predict the competency gaps in the high-level personnel of a project [30].

In the project quality management, Shi (2009) proposed a model as a combination of an unascertained system and an artificial neural network to evaluate the quality of construction projects [52]. The author mentioned that an artificial neural network deployed for processing nonlinear problems and an unascertained system handled uncertain knowledge by copying the human brain's thinking rational [52].

Of 58, 11 papers focused on the project schedule management. Gersmann & Hammer (2005) investigated the resource constraint project scheduling problem (RCPSP) by analyzing restrictions on the availability of the project resources [60]. The authors adapted a simple greedy strategy for solving the problem [60]. Masmoudi & HaïT (2013) integrated fuzzy parameters in two project scheduling techniques of resource leveling and resource-constrained scheduling to develop the fuzzy resource leveling problem and fuzzy resource-constrained project scheduling problem [61]. Subsequently, the authors solved the problems by a genetic algorithm and a greedy algorithm [61]. Podolski (2017) investigated an optimum management of the resources in the project scheduling by deploying a metaheuristic tabu search algorithm. The author indicated that, by using the proposed model, the project execution time can reduce more than 50% [53].

Lu (2002) used a neural network -based approach to determine parameters of the beta distributions to enhance project evaluation and review technique (PERT) simulation [62]. Cheng & Hoang (2018) tried to estimate the project duration at the planning phase by developing an AI approach named firefly-tuned least squares support vector machine [63]. The proposed method is a fusion of (1) a least squares support vector machine (LS-SVM), to generalize input-output mapping, and (2) a firefly algorithm, to discover a suitable set of the tuning parameters of LS-SVM [63]. Gondia and associates (2019) used a decision tree and a naïve Bayesian classification to predict essential risk factors affecting the projects' delay [64]. The authors reported that the naïve Bayesian classification outperformed the decision tree in terms of the prediction performance [64]. One study developed a neural network–long short-term memory model to predict the estimated schedule to completion (ESTC) [65].

Yang & Wang (2008) developed an online planning recommendation system based on a revised case-based reasoning algorithm to help project managers in developing the project plans and knowledge discovery [68]. Kartelj and associates (2014) developed an EM K-NN algorithm as a fusion of electromagnetism-like algorithm (EM), metaheuristic optimization technique, and k-nearest neighbors algorithm for knowledge discovery in the construction management [67].

Wauters & Vanhoucke (2017) forecasted the project duration by using a nearest-neighbor based extension model [69]. Another study used five artificial intelligence methods including decision tree, bagging, random forest, boosting, and support vector machine to forecast the final duration of a project and were gauged against the best performing earned value management/earned schedule (EVM/ES) methods [7]. The authors reported that AI methods outperformed the EVM/ES methods if the test and training sets are similar to one another [7]. Relich & Muszyński (2014) used artificial neural networks and a fuzzy neural system to predict the duration of a project [66]. The sankey diagram of the application of different AI techniques in the project schedule management is represented in Figure 4.

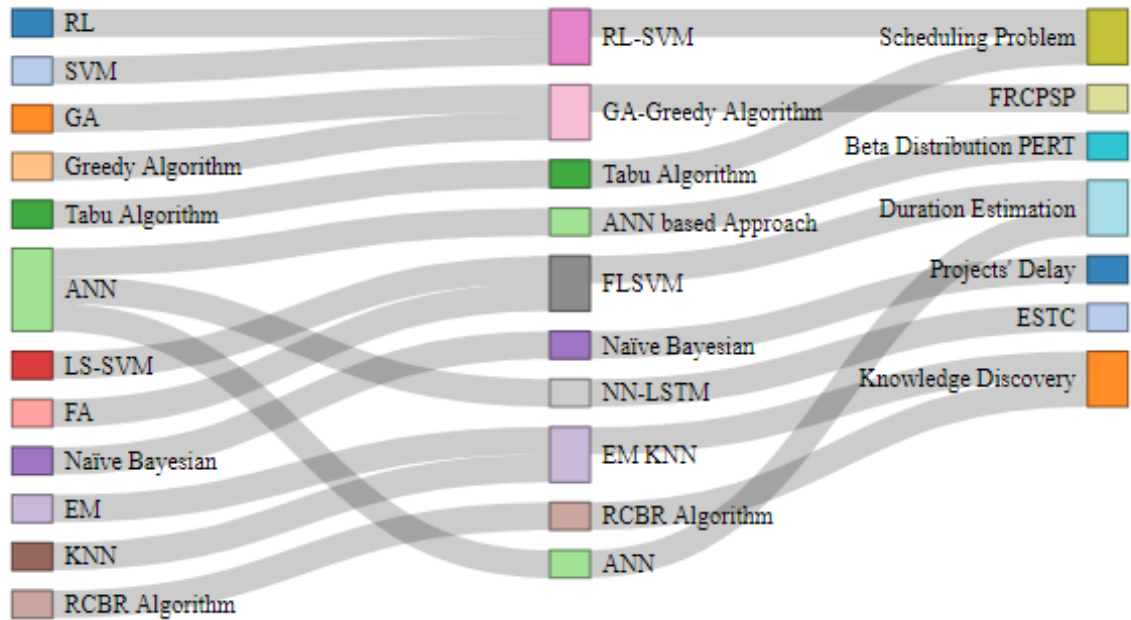

Figure 4. Sankey diagram among single algorithms (left part), final models (middle part) and process of project schedule management (right part)

4 papers out of 58 papers focused on project risk management. One study investigated the risk of requirements as a part of the risk assessment problem by using different machine learning techniques such as Bayesian networks classifiers, Naïve Bayes, k-nearest neighbors and J48 decision tree [55]. The authors reported that Bayesians networks outperformed other techniques [55]. Hu and associates (2013) developed a model using Bayesian networks with causality constraints to investigate a correlation between project outcome and risk factors and compared with other algorithms including general Bayesian networks, C4.5, logistic regression, and Naïve Bayes [56].

In terms of the project scope management, Perini and associates (2012) studied the requirements prioritization by developing a Case-Based Ranking method [58]. Requirements prioritization is about selecting the most important requirements in the processes of software development [58].

# 5. Conclusion

This literature review analyzed published papers from 2000 to 2019 that focused on the application of AI in PM. Included papers were reviewed to identify AI models and corresponding PM processes. The result indicated that support vector machine, neural networks, and genetic algorithms are widely applied to different processes of PM. Common PM processes among included papers are including effort predictions, cost estimation, and project success factors. Even though AI techniques have not been used in multiple PM processes, the included papers demonstrated the efficiency of AI models in several PM processes.

Two main processes in the initiating process group are developing a project charter and identifying stakeholders. Published papers in the initiating process group are focused on the aspects of the effort estimation, determining project success factors, and identifying stakeholders. As it is represented in Table 3, in the project integration management, 15 papers studied the effort estimation and project success factors and 6 papers focused on decision making process and dispute problems.

Table 3. Density map of included papers based on PM knowledge areas and process groups

| knowledge Areas | Project Management Process Groups | | | | |
|---|---|---|---|---|---|
| | Initiating | Planning | Executing | Monitoring and Controlling | Closing |
| **Project Integration Management** | 15 papers in prediction of effort and success factors | No paper | No paper | 6 papers in decision making and dispute problems | No Paper |
| **Project Scope Management** | | 2 papers in collection of requirement | | No paper | |
| **Project Schedule Management** | | 8 papers in developing schedule | | 3 papers in schedule control | |
| **Project Cost Management** | | 8 papers in cost estimation | | 8 papers in cost control | |
| **Project Quality Management** | | No paper | No paper | 1 paper in quality control | |
| **Project Resource Management** | | No paper | 1 paper in team development | 1 paper in resource control | |
| **Project Communications Management** | | No paper | 1 paper in communication management | No paper | |
| **Project Risk Management** | | 6 papers in risk planning | No paper | No paper | |
| **Project Procurement Management** | | No paper | No paper | No paper | |
| **Project Stakeholder Management** | No paper | No paper | No paper | No paper | |

The most popular process group is planning and 24 papers studied the application of AI in different processes of the planning process group. AI has been used in developing schedules, determining budget and cost, planning risk, and collecting requirements. However, there is no paper in the application of AI in the processes of planning quality, estimating resources, planning communication, and planning procurement. Because planning processes are interrelated, it is important to investigate the application of AI in all aspects of planning.

Nineteen papers focused on different processes of monitoring and controlling process group. Facilitating decision making and dispute problems, controlling costs, controlling duration, controlling quality, and controlling resources are investigated by included papers. However, there is no paper in the processes of monitoring communication, monitoring risks, controlling procurements, and monitoring stakeholder engagement.

Closing and executing are the least popular process groups among included papers. In the closing process group, no paper investigate the application of AI in the closing project and collecting lessons learned. In the executing process group, only two papers investigated the processes of developing teams and managing communication. Accordingly, predicting competence gaps in high-level personnel in project management and managing project documents are investigated in this process group. Processes of managing quality, risk responses, conducting procurements, and managing stakeholder engagement have not been studied.

Author Contributions: Mohammad Reza Davahli: Methodology, Writing - Original Draft and Revisions;

Conflicts of Interest: The author declare no conflict of interest.